\documentclass[letterpaper, 10 pt, conference]{ieeeconf}  % Comment this line out
\IEEEoverridecommandlockouts                              % This command is only
\usepackage{graphicx} % for pdf, bitmapped graphics files
\usepackage{bm}

\usepackage{epsfig} % for postscript graphics files
\usepackage{mathptmx} % assumes new font selection scheme installed
\usepackage{times} % assumes new font selection scheme installed
\usepackage{amsmath} % assumes amsmath package installed
\usepackage{amssymb}  % assumes amsmath package installed

\usepackage{color}

\title{\LARGE \bf
Pneumatic bladder links with wide range of motion joints for articulated inflatable robots
}

\author{Katsu Uchiyama and Ryuma Niiyama% <-this % stops a space
\thanks{}% <-this % stops a space
\thanks{The authors are with the School of Science and Technology, Meiji University, 1-1-1 Higashi Mita, Tama-ku, Kawasaki-shi, Kanagawa}%
}

\begin{document}

\maketitle
\thispagestyle{empty}
\pagestyle{empty}

%%%%%%%%%%%%%%%%%%%%%%%%%%%%%%%%%%%%%%%%%%%%%%%%%%%%%%%%%%%%%%%%%%%%%%%%%%%%%%%%
\begin{abstract}
Exploration of various applications is the frontier of research on inflatable robots. We proposed an articulated robots consisting of multiple pneumatic bladder links connected by rolling contact joints called Hillberry joints. The bladder link is made of a double-layered structure of tarpaulin sheet and polyurethane sheet, which is both airtight and flexible in shape. The integration of the Hilberry joint into an inflatable robot is also a new approach. The rolling contact joint allows wide range of motion of $\pm 150 ^{\circ}$, the largest among the conventional inflatable joints. Using the proposed mechanism for inflatable robots, we demonstrated moving a 500 g payload with a 3-DoF arm and lifting 3.4 kg and 5 kg payloads with 2-DoF and 1-DoF arms, respectively. We also experimented with a single 3-DoF inflatable leg attached to a dolly to show that the proposed structure worked for legged locomotion.
\end{abstract}

%%%%%%%%%%%%%%%%%%%%%%%%%%%%%%%%%%%%%%%%%%%%%%%%%%%%%%%%%%%%%%%%%%%%%%%%%%%%%%%%
\section{INTRODUCTION}
%%% このパラグラフは消してもいいかもしれない。
Inflatable robots are robots in which inflatable structures are used for main body. Typical characteristics include light weight, storage efficiency, compliance, low cost, and softness when the internal air pressure is low. Considering these advantages, potential applications for inflatable robots would be robots that can travel and manipulate objects in outer space, at disaster sites, on uneven terrain, and inside buildings.

An important factor in constructing an inflatable robot is how the joint structure is made. Most inflatable robots' links are made up of a tubular membrane that is inflated with air. On the other hand, joints are implemented in a wide variety of ways, whether they are made of membranes alone~\cite{Sanan2011:Inflatable_Robot_Arm} or rigid materials such as metal or 3D printed parts~\cite{Kim2018:High_Accuracy_Inflatable_Robot}, and how they are structured.

Inflatable robots have a lower density of structural elements compared to robots consisting of rigid structures. Therefore, ingenuity is required to increase the degree of freedom of the robot arm. One approach is a multi-link robot arm that cancels its own weight by using helium gas-filled inflatable links and floats in the air~\cite{Takeichi2017:20m_Giacometti_Arm},~\cite{Li2023:Large_Scale_IRA}. Each joint of these floating robot arms has one degree of freedom, and many links are connected to form a hyper-redundant robot arm. These robotic arms are intended for inspection purposes and have almost zero payload. There is a study of an inflatable robotic arm with a relatively large payload capacity that uses multiple active chambers between metal plates to form a two degrees-of-freedom joint~\cite{Oh2023:Large_Scale_Inflatable_Robotic_Arms}. However, the feature that the entire structure is made of soft materials is lost in exchange for the stiffness of the joint.

In this research, we consider increasing the range of motion of each joint part as a strategy not only to increase the degree of freedom but also to expand the working space. In addition, the aim is to avoid utilizing rigid structures at joints, while still taking advantage of the features of inflatable structures. Since the joint structure and actuation of an inflatable robot are closely related, it is necessary to consider the actuation method as much as the joint structure to achieve these goals.

\begin{figure}
    \centering
    \includegraphics[width=0.40\textwidth]{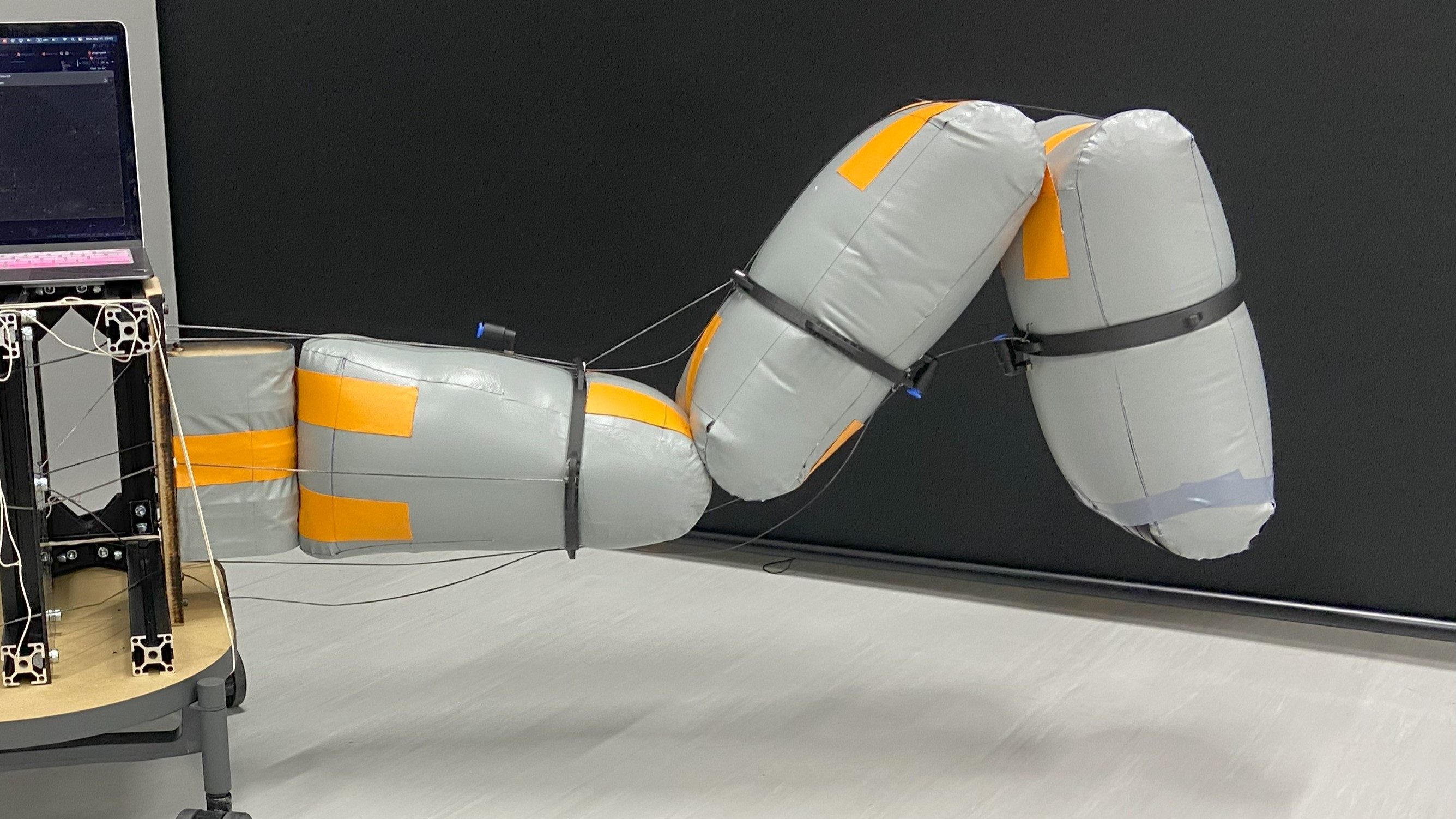}
    \caption{A 3-DoF inflatable robot with Pneumatic bladder link}
    \label{fig:arm_3dof}
\end{figure}

We focus on robots that do not use rigid structures for joints. Inflatable robot using electric actuators are often used for interaction purposes and have advantages in terms of ease of handling, and can avoid using hard structures at joints by using wire tendon mechanisms~\cite{Sanan2011:Inflatable_Robot_Arm},~\cite{Qi2017:Safe_Inflatable_Robot_for_HRI},~\cite{Seong2019:Low_pressure_Soft_Inflatable_Joint}. Even if the size of the robot is human scale or larger and the pulling length increases, it can be handled by increasing the pulling length and increasing the stroke as a wire drive. This would be compatible with the larger size-enabling characteristics of inflatable structures.

Many inflatable robots driven by pneumatic bag actuators have relatively high payloads. Bodily \textit{et al.} developed an inflatable robotic arm that could be attached to a space rover and lift falling rocks and other objects, although specific figures were not available~\cite{Bodily2017:Multi_objective_Design_Optimization}. Liang \textit{et al.} showed that a 2-DoF SRA can lift a 5 kg payload~\cite{Liang2018:Soft_Robotic_Arm}. Oh \textit{et al.} lifted a 3 kg payload in a total of 6-DoF inflatable robotic arm~\cite{Oh2023:Large_Scale_Inflatable_Robotic_Arms}. Some have been proposed that can output large forces and have high positional accuracy with system integration~\cite{Hyatt2019:Configuration_Estimation_IRA}. While the specific weight is unknown, it does accomplish the task of picking up the stone and placing it in the box.

However, due to the structure of how these pneumatic bag actuators are attached, the range of motion of the robot itself is thought to be limited to a certain extent, even if the actuator itself has a wide range of motion. On the other hand, robot using pneumatic artificial muscles (PAM)~\cite{Li2020:Design_Modeling_Characterization_IRA},~\cite{Takeichi2017:Giacometti_Arm} have advantages such as light weight. However, joints driven by PAMs are expected to be incompatible due to the narrow stroke of the actuator, resulting in a narrow range of motion. From these, in order to obtain a wide range of motion, it is considered suitable to use a tendon drive with an reel of wire driven by electric motor for the actuation.

In this work, we fabricate 1-DoF to 3-DoF inflatable robots (Fig.\ref{fig:arm_3dof}) with tendon-driven rolling contact joints that allow a wide range of motion in an inflatable structure. In addition, for future expandability of the robot, the links is be fabricated separately at the link level to treat them as units or modules. The wide range of motion joint and the modular and expandable links, which had not been realised for inflatable robots, enable the production of robot arms with a wide working space with various degrees of freedom by means of combinations. We perform basic verification and weight lifting with the robot arms. We also conducted locomotion experiment with a single inflatable leg on a dolly.

\section{Method}

\subsection{Joint Design}
\begin{figure}
    \begin{minipage}[b]{0.49\columnwidth}
        \centering
        \includegraphics[width=0.99\textwidth]{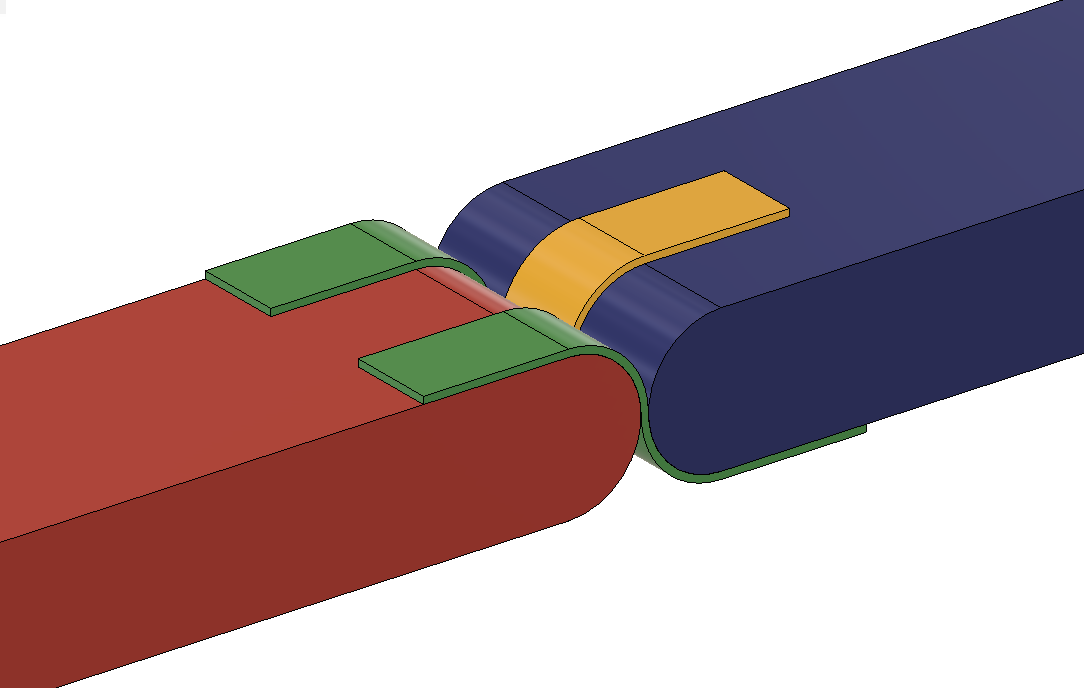}
        \caption{Model of Hilberry joint}
        \label{fig:hilberry_model}
    \end{minipage}
    \begin{minipage}[b]{0.49\columnwidth}
        \centering
    \includegraphics[width=0.99\textwidth]{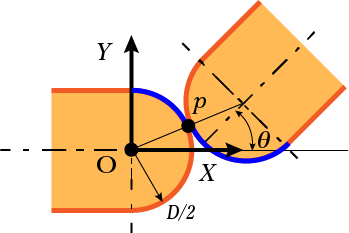}
    \caption{Coordinate system for the center of rotation of the Hilberry joint}
    \label{fig:joint_rot_center}
    \end{minipage}
\end{figure}

As a form of joint with a wide range of motion, we focus on the Hilberry joint, which is a type of rolling contact joint proposed by Hilberry \textit{et al.}~\cite{Hilberry1973:Rolling_Contact_Joint}. The model of this joint is shown in Fig.\ref{fig:hilberry_model}. The two links are constrained by attaching ribbon-like straps (shown in Fig. \ref{fig:hilberry_model} as yellow or green color objects) wrapped around both cylinder shaped link ends. When the joint rotates, the cylindrical part of the link rolls and moves. Therefore, the geometric feature is that the center of rotation changes with the angle, and the straight line where the two links tangent is the center of rotation at that angle.

The length of the strap attached is given as the sum of the front and back tangents of the two links as follows, so the strap length is constant and acts as a constraint. The path of the strap is shown by the blue curve in Fig.\ref{fig:joint_rot_center}.
\begin{equation}
    S = \pi D \dfrac{90^\circ - \theta / 2}{360^\circ} + \pi D \dfrac{90^\circ + \theta / 2}{360^\circ} = \dfrac{\pi D}{2}
\end{equation}

If the coordinate system is set up as in Fig.\ref{fig:joint_rot_center}, the position of the center of rotation is given as follows.
\begin{equation}
    \bm{x} = \begin{bmatrix}
        \dfrac{1}{2} D \cos \dfrac{1}{2} \theta & \dfrac{1}{2} D \sin \dfrac{1}{2} \theta
    \end{bmatrix}^{\rm{T}}
\end{equation}

When this joint is used in an inflatable structure, the straps for constraint can be made of the same material as the membrane for the structure. The fact that the material of the link itself and the material for the constraint are the same means that there is no need to use a rigid structure for the joint, and the affinity between the inflatable robot and the Hilberry joint is very high. This does not affect the advantages of the inflatable structure, such as light weight and high storage efficiency, and does not impair the property of the structure.

\subsection{Pneumatic Bladder Link Design}
Make it possible to treat the links as a unit, with a view to treating it as modular in order to increase the degree of freedom by connecting multiple links. In other words, as an inflatable robot, one pneumatic bladder per link would be used. As an inflatable robot, it will be made with a separate bag per link. The combination of pneumatic bladder links can be selected according to the desired degree of freedom and arrangement.

To make a Hilberry joint, the ends of adjacent links must be cylindrical like the red and blue links in Fig.\ref{fig:hilberry_model}, respectively. To reduce deflection in a direction perpendicular to the direction of rotation, the contact surfaces of the joints are long in a direction parallel to the axis of rotation.

In order to expand the range of motion of the arm as a whole, links with parallel and orthogonal relationships between the cylindrical sides of the two ends were fabricated. This allows for 3 dimensional motion depending on the combination of the links.

\subsection{Fabrication}
The size of the pneumatic bladder link was made to be easy to fabricate and is about the size of a medium-sized robotics arm. The minimum possible curvature of the link cross section was set at about 0.05~mm in order to maintain strength and to prevent wrinkling and buckling of the membrane, and the internal pressure was estimated to be 100~kPa based on the relationship between the section modulus of the thin-walled structure and the axial stress.

The outer membrane of the Pneumatic Bladder Link is made of 0.35~mm thick tarpaulin sheet. This material has low elasticity, withstands high pressure, and has excellent availability. We determined the Young's modulus of the tarpaulin. From this experiment, the value was found to be 320 $\sim$ 560~MPa. According to formula for deriving elongation from pressure, Young's modulus, radius and thickness in thin-walled structures, $\delta r = p r^2 / Et $, the radial elongation of the membrane when sealed is estimated to be around $\delta r = 0.223~\rm{mm}$. This amount of elongation is not different from the error in manufacture and is therefore considered to have little impact.

\begin{figure}
    \centering
    \includegraphics[width=0.40\textwidth]{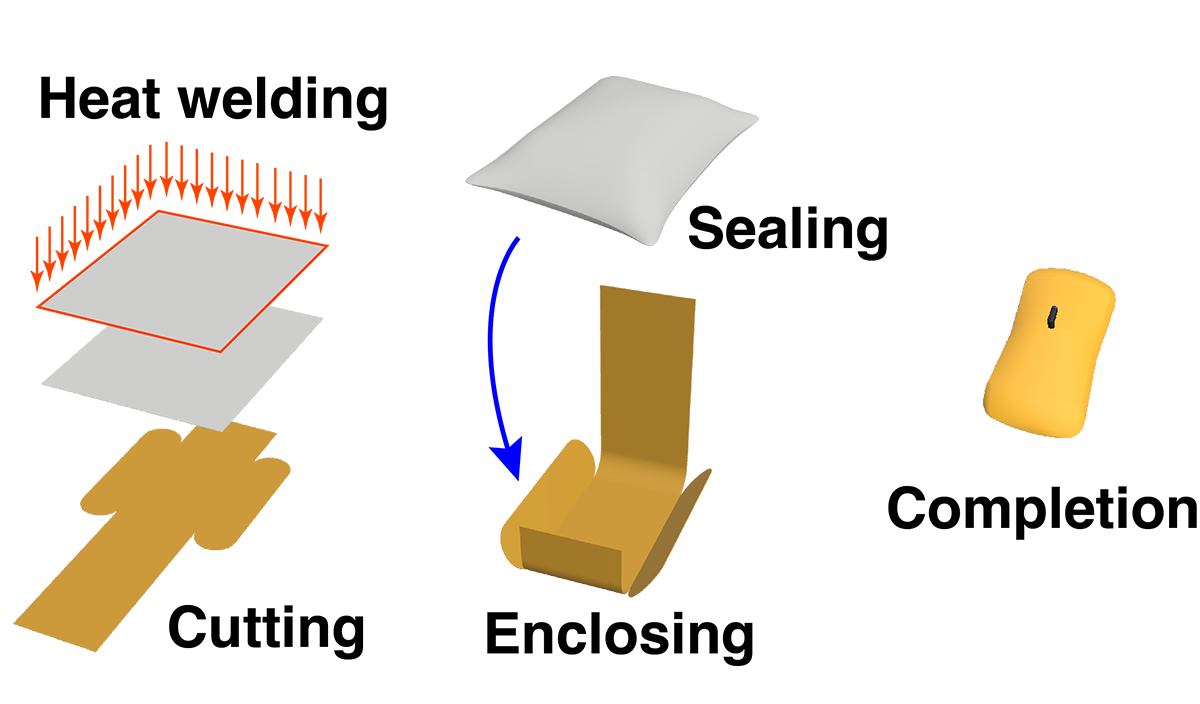}
    \caption{Production process of pneumatic bladder link}
    \label{fig:inflatable_link_production}
\end{figure}

\begin{figure}
    \centering
    \includegraphics[width=0.45\textwidth]{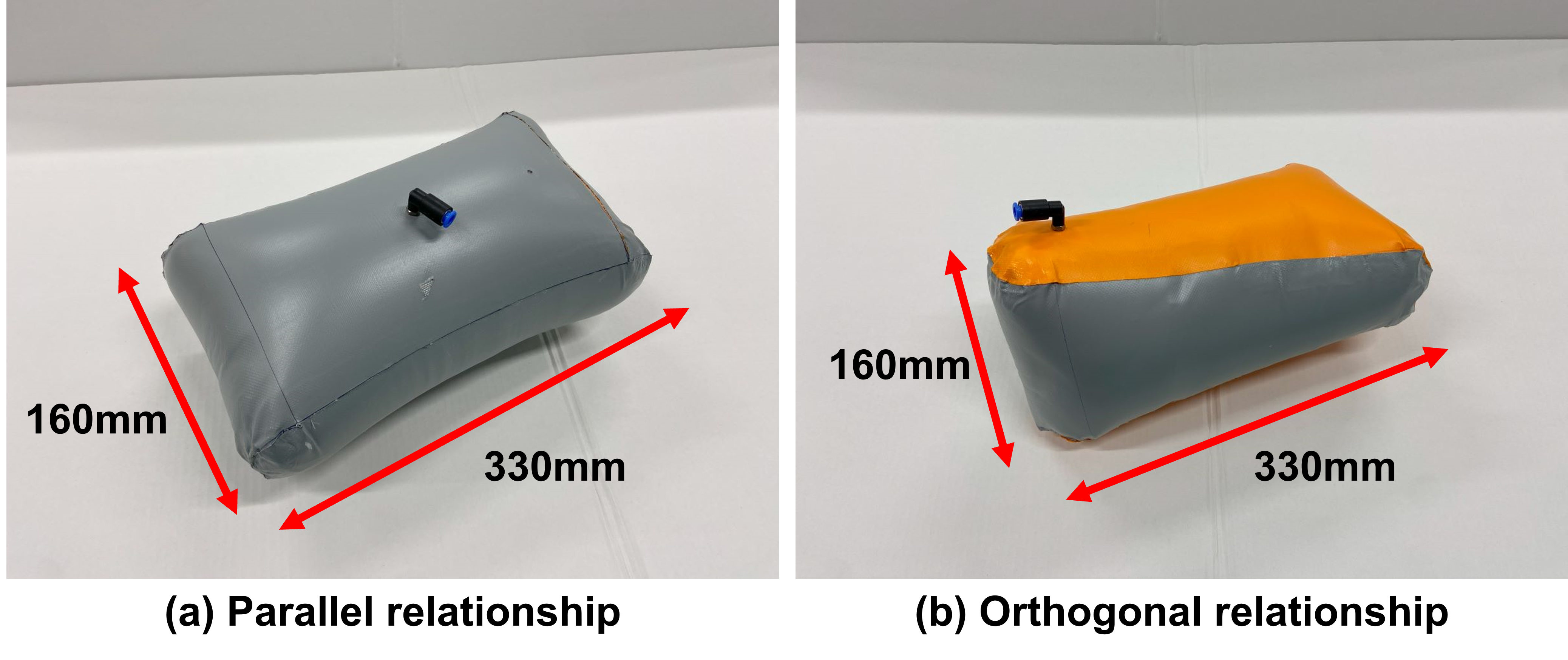}
    \caption{Fabricated Pneumatic bladder link}
    \label{fig:pneumatic_bladder_links}
\end{figure}

It was difficult to achieve sufficient airtightness with only a tarpaulin sheet. Therefore, a polyurethane sheet was installed inside for manufacturing convenience. Although the polyurethane sheet is elastic, its shape follows the outer membrane by tarpaulin sheet because it fits inside the bladder by the tarpaulin. Therefore, the shape of the links is ensured by the tarpaulin sheet, and the air tightness is ensured by the polyurethane sheet. As shown in Fig.\ref{fig:inflatable_link_production}, two polyurethane sheets were overlapped and laminated on four sides with heat sealer, sealed in a tarpaulin sheet cut to the desired shape, and the tarpaulin sheet was closed with adhesive. A tube coupler (SMC KCL06-M5) was attached for air supply. Shows the pneumatic bladder link produced for Fig.\ref{fig:pneumatic_bladder_links}.

To constrain the pneumatic bladder link, a tarpaulin sheet was cut into a rectangular shape and attached like the orange colored strap in Fig.\ref{fig:arm_3dof}. By repeating this process according to the desired degree of freedom, a series of robot arms are completed.

We fabricated an Anchor Point Ring to attach the tendon drive string. The length of one round is made slightly smaller than the circumference of the pneumatic bladder link, so that it is fixed from bulging when supplied air. When the string passes through to drive the next and subsequent links in the chain, an eyelet guide is installed to reduce damage to the string due to wear. In this case, we used a servo motor (Dynamixel XM430-W350-R) for pulling the tendon drive, and Dynamixel's U2D2 Power Hub Board to control the servo motor. And we use 1mm diameter UHMWPE (Ultra High Molecular Weight Polyethylene) cord as the tendon wire for the servo motor.

\begin{figure}
    \centering
    \includegraphics[width=0.35\textwidth]{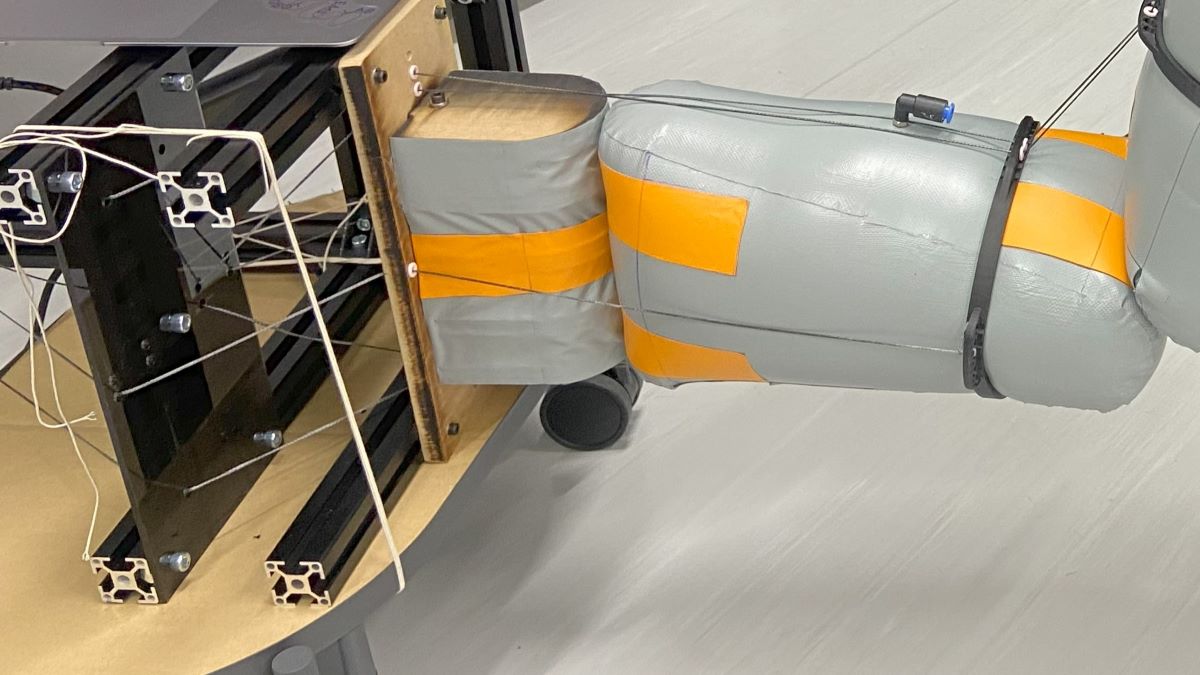}
    \caption{Mounting part of robot arm and trestle}
    \label{fig:fix_to_base_imposter}
\end{figure}

For the operation, a trestle was fabricated to fix the robot arm. The base of the robotic arm is fixed to a part with cylindrical sides made of a rigid structure as shown in Fig.\ref{fig:fix_to_base_imposter}. Since the structure of this cylindrical side is similar to the end of the pneumatic bladder link, they are connected as Hilberry joints.

\begin{figure}
    \centering
    \includegraphics[width=0.48\textwidth]{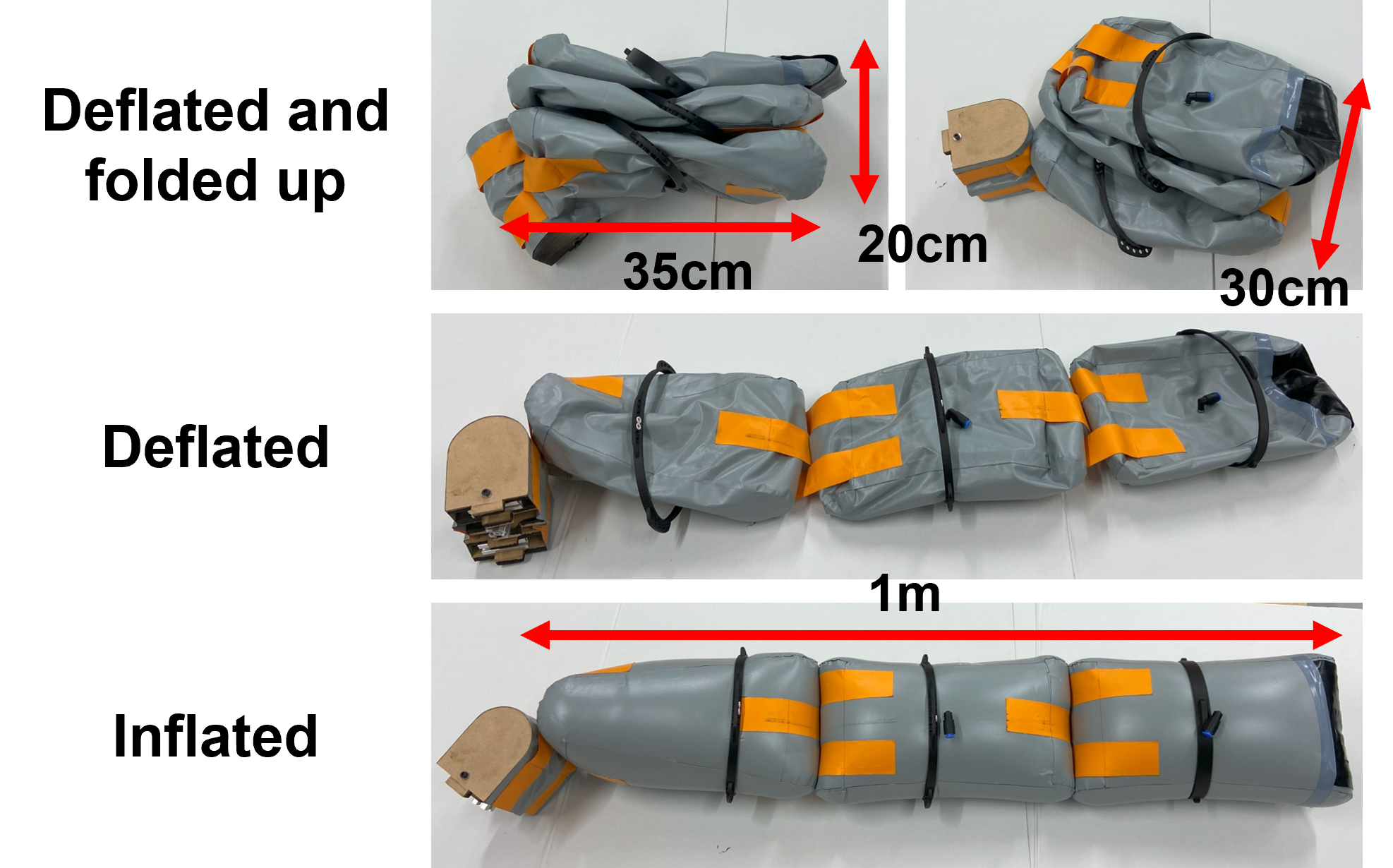}
    \caption{3-DoF robotic arm from folded to inflated state with air removed. The dimensions of each link after inflation follow Table.\ref{table:params}. Since these are the dimensions in their natural state, they become smaller when folded by applying force.}
    \label{fig:inflated_deflated_folded}
\end{figure}

A comparison between the uninflated and inflated robotic arm is shown in Fig.\ref{fig:inflated_deflated_folded}
    . The bulging creates tension in the joint restraint straps and frictional forces with the links, making the joint areas tighter.

The physical parameters of the fabricated robot arm are shown in Table.\ref{table:params}. Unless otherwise noted in this study, the robot is configured with these parameters.

\renewcommand{\arraystretch}{1.5}
\begin{table}[h]
 \caption{Physical parameter}
 \label{table:params}
 \centering
  \begin{tabular}{lr}
   \hline
   Items & Value \\
   \hline
    Mass (1 link) & 0.15 kg \\
    Internal air pressure & 100 kPa \\
    Link length & 330 mm \\
    Link width & 160 mm \\
    Maximum link height (100 kPa) & 160 mm \\
    Diameter of joint & 80 mm \\
    \hline
    Range of motion (100 kPa) & $-$150 deg -- 150 deg \\
   \hline
  \end{tabular}
\end{table}
\renewcommand{\arraystretch}{1}

\section{Modeling}

\subsection{Moment Arm and Torque}
\begin{figure}
    \centering
    \includegraphics[width=0.40\textwidth]{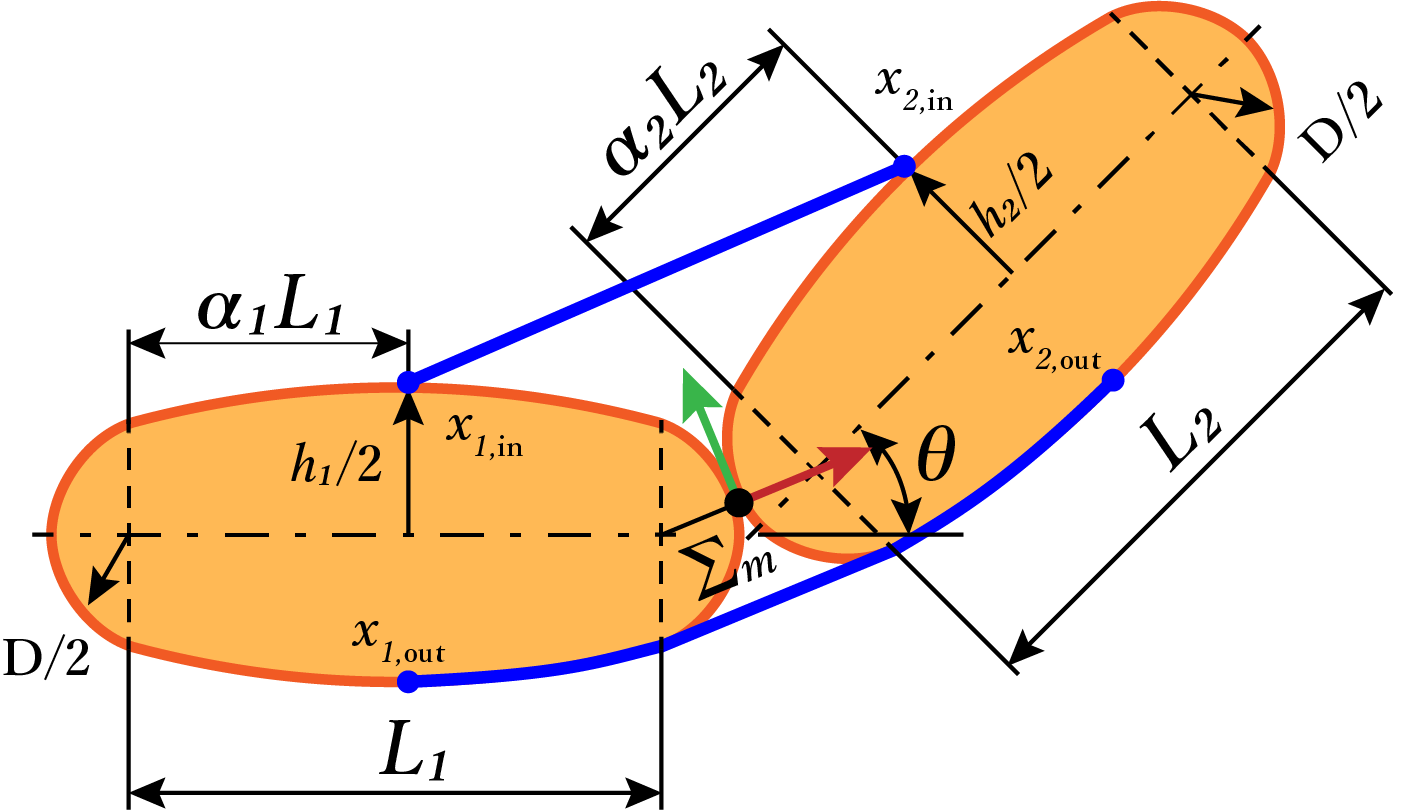}
    \caption{Geometry such as the path of a string between two links}
    \label{fig:moment_arm}
\end{figure}

The moment arm of this form of tendon drive depends on the angle of rotation. It is also necessary to consider that the center of rotation varies with the angle. The center of rotation of the Hilberry Joint is the origin, as shown in Fig.\ref{fig:moment_arm}. Let $D_i$ be the diameter of the cylinder side part of the link end, $L_i$ be the link length excluding the link end, and $h_i$ be the height considering the bulge in the center of the link. In this case, the positions $\bm{x}_{\rm{A},1}$ and $\bm{x}_{\rm{A},2}$ in the $\Sigma_{m}$ coordinate system of the Anchor Point Ring of the front and rear links are given as follows.
\begin{equation}
    \bm{x}_{\rm{A},1} =
        \begin{bmatrix}
            -(1 - \alpha_1) L_1 \cos \dfrac{1}{2} \theta - \dfrac{1}{2} D + \dfrac{1}{2} h_1 \sin \dfrac{1}{2}\theta \\[10pt]
            -(1 - \alpha_1) L_1 \sin \dfrac{1}{2} \theta + \dfrac{1}{2} h_1 \cos \dfrac{1}{2}\theta
        \end{bmatrix}
\end{equation}
\begin{equation}
    \bm{x}_{\rm{A},2} =
        \begin{bmatrix}
            \alpha_2 L_2 \cos \dfrac{1}{2} \theta + \dfrac{1}{2} D - \dfrac{1}{2} h_2 \sin \dfrac{1}{2}\theta \\[10pt]
            \alpha_2 L_2 \sin \dfrac{1}{2} \theta + \dfrac{1}{2} h_2 \cos \dfrac{1}{2}\theta
        \end{bmatrix}
\end{equation}
Since the moment arm at the inner side is the distance from the origin, it is given as follows.
\begin{equation}
    r = \frac{|\bm{x}_{\rm{A},1} \times \bm{x}_{\rm{A},2}|}{\|\bm{x}_{\rm{A},1} - \bm{x}_{\rm{A},2}\|}
\end{equation}
In particular, if links of the same shape are used at the front and rear and an Anchor Point Ring is attached in the middle, $\alpha_1 = \alpha_2 = 1/2$, $L_1 = L_2 = L$, $D_1 = D_2 = D$, $h_1 = h_2 = h$ is given. The moment arm can be given by finding the line connecting the two points and the distance from the origin.
\begin{equation}
    r = \frac{1}{2} \left( L \sin \frac{1}{2} \theta + h \cos \frac{1}{2} \theta \right)
\end{equation}
The moment arm at the outer side has a string passing along the outside of the link. It is given as follows.
\begin{equation}
    r = \dfrac{1}{2} \left( h + D \sin \dfrac{1}{2} \theta \right)
\end{equation}
From the moment arm, the torque $T$ when force $f$ is applied is given as follows.
\begin{equation}
    T = rf = \left\{\begin{array}{l}
        \dfrac{1}{2} \left( L \sin \dfrac{1}{2} \theta + h \cos \dfrac{1}{2} \theta \right) f, \rm{inner} \\[5mm]
        \dfrac{1}{2} \left( h + D \sin \dfrac{1}{2} \theta \right) f, \rm{outer}
    \end{array}
\right.
\end{equation}

\begin{figure}
    \centering
    \includegraphics[width=0.22\textwidth]{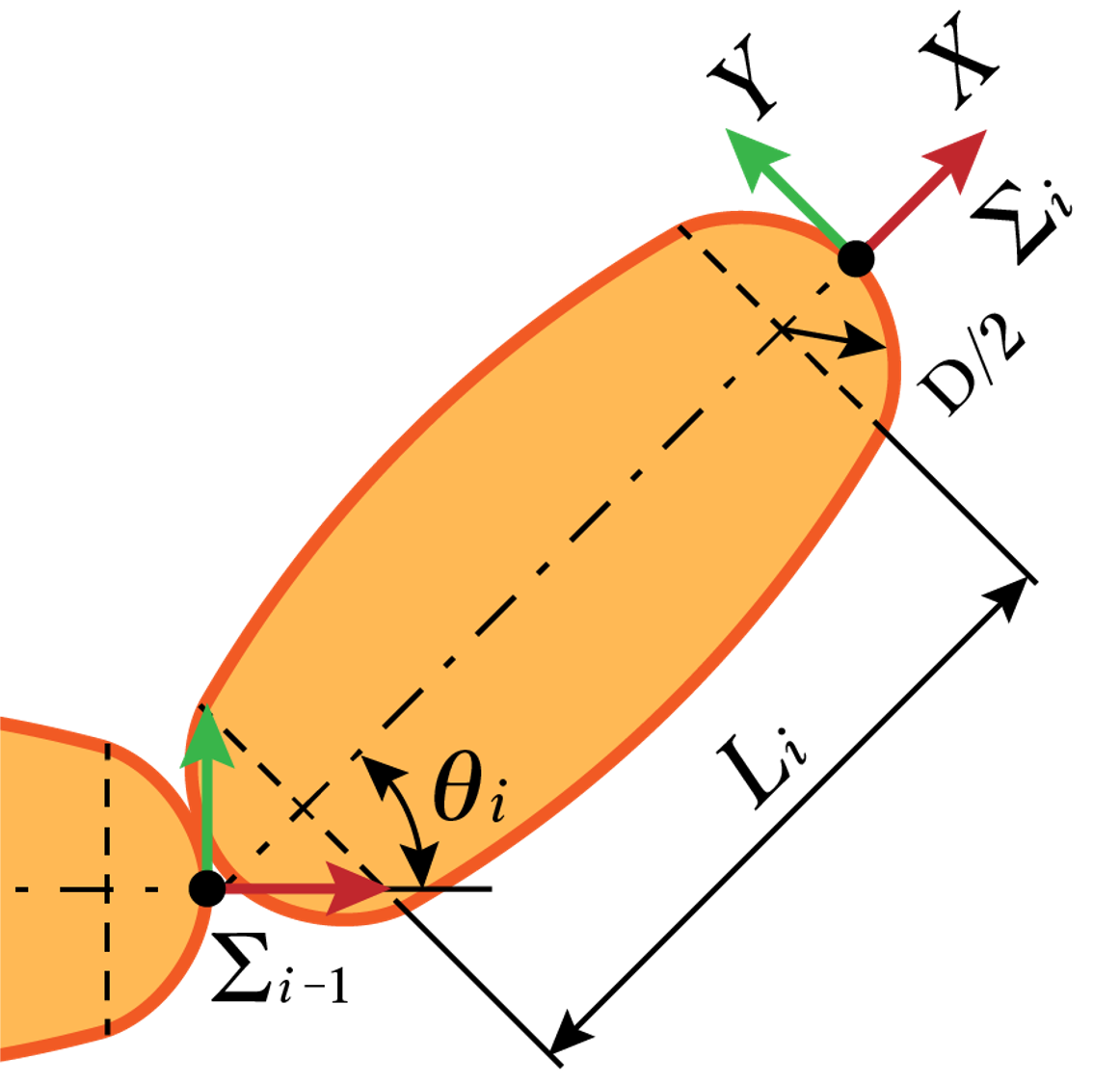}
    \caption{Coordinate system for each link}
    \label{fig:coordinate_system}
\end{figure}

\subsection{Kinematics}
Consider the kinematics to describe the position of the tip of each link. The coordinate system is set up as shown in Fig.\ref{fig:coordinate_system}. Take a coordinate system with the origin at the tip of the link, as in link $i$. The positional relationship of each link coordinate system is expressed as follows.
\begin{equation}
    \bm{x}_i = \begin{bmatrix}
        -\dfrac{1}{2} D + \left( L_i + \dfrac{1}{2} D \right) \cos \theta_i + D \cos \dfrac{1}{2} \theta_i \\
        \left( L_i + \dfrac{1}{2} D \right) \sin \theta_i + D \sin \dfrac{1}{2} \theta_i \\
        0
    \end{bmatrix}
\end{equation}

If the rotation axes of the joints at both ends of the link are in a parallel relationship, the homogeneous transformation matrix is given as follows.
\begin{equation}
    {}^{i-1}T_{i} = \begin{bmatrix}
        R_z(\theta_i) & \bm{x}_i \\[3mm]
        \bm{0} & 1
    \end{bmatrix}
\end{equation}
In the same way, in the case of an orthogonal relationship, it is given as follows.
\begin{equation}
    {}^{i-1}T_{i} = \begin{bmatrix}
        R_z(\theta_i)R_x(90^{\circ}) & \bm{x}_i \\[3mm]
        \bm{0} & 1
    \end{bmatrix}
\end{equation}

\section{Experiments}
    
\subsection{Motion Capture}
\begin{figure}
    \centering
    \includegraphics[width=0.48\textwidth]{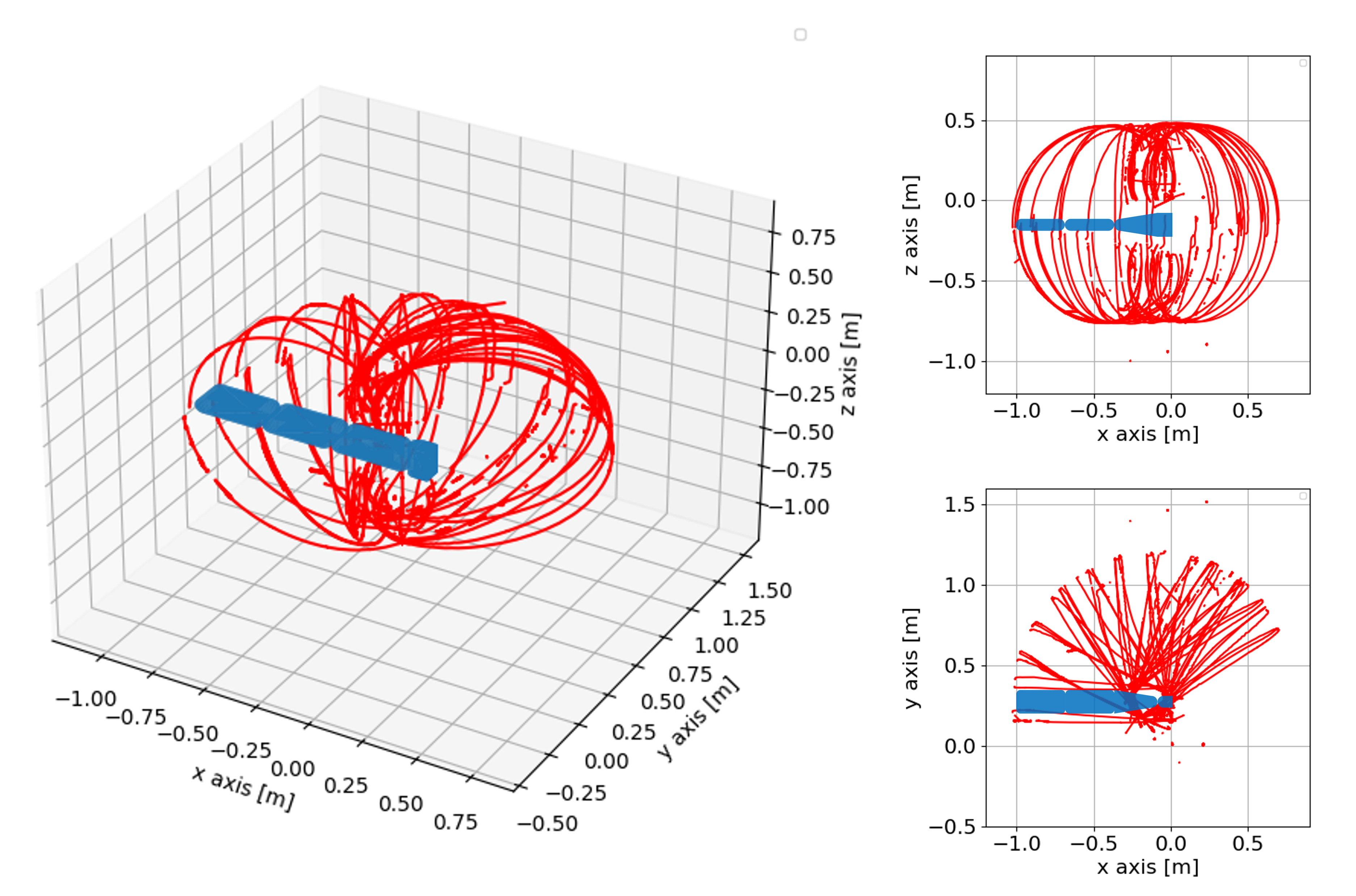}
    \caption{Result of motion capture. The blue model illustrates the initial position of the robot arm. The view from the side is plotted in the upper right and the view from directly above is plotted in the lower right.}
    \label{fig:mocap_3dof}
\end{figure}

Motion capture of a 3-DoF robot arm was performed using OptiTrack's Prime 13W. The result of drawing the workspace is shown in Fig.\ref{fig:mocap_3dof}. Since the robot is a line symmetric robot, half of the trajectory is shown. As a countermeasure against the markers being obscured by blind spots, we mounted the motion capture markers in two locations on either side of the cylindrical side of the 3-DoF arm's end link.

In this experiment, Joint1 was moved only $+150^{\circ}$, and Joint2 was moved $\pm 150^{\circ}$; Joint3 was moved to the extent that it did not interfere with the link, trestle, desk, etc., after the maximum bending of Joint2.

\subsection{Lifting}
\begin{figure}
    \centering
    \includegraphics[width=0.48\textwidth]{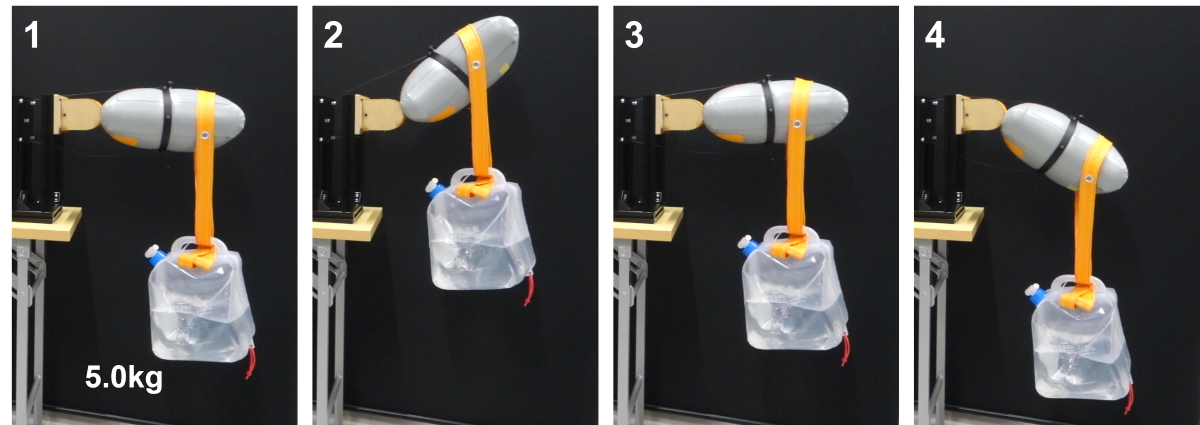}
    \caption{Payload Testing in 1-DoF. Weight: 5 kg. Moment arm: 60 cm}
    \label{fig:demo_1dof_5kg_lift_up}
\end{figure}

\begin{figure}
    \centering
    \includegraphics[width=0.40\textwidth]{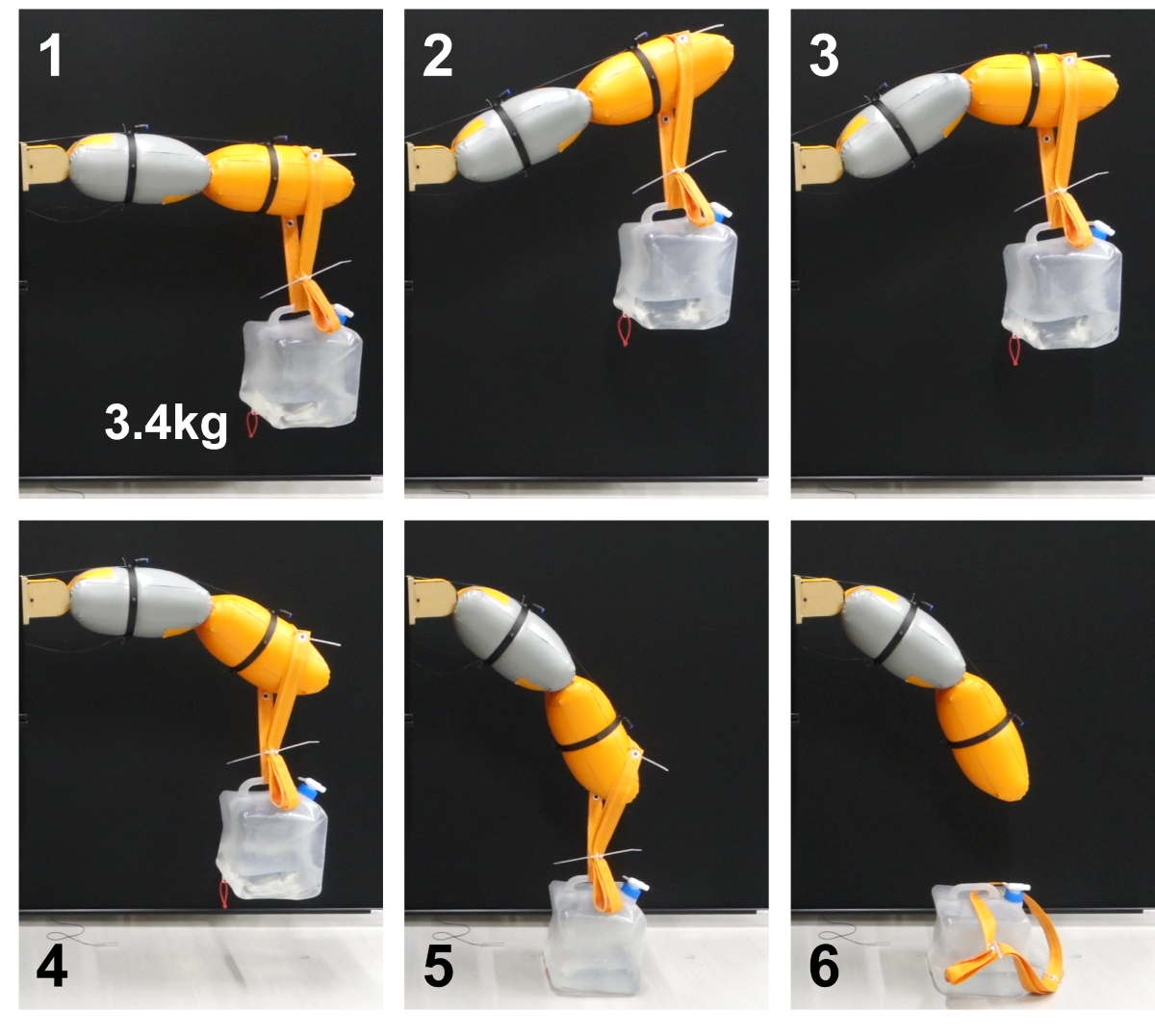}
    \caption{Payload Testing in 2-DoF. Weight: 3.4 kg. Moment arm: 25 cm}
    \label{fig:demo_2dof_3_4kg_lift_up}
\end{figure}

We performed weight lifting experiments on the 1-DoF and 2-DoF robotic arms. 5.0~kg weight was placed 25~cm from the base of the 1-DoF arm in Fig.\ref{fig:demo_1dof_5kg_lift_up} and 3.4~kg weight was placed 60~cm from the base of the 2-DoF arm in Fig.\ref{fig:demo_2dof_3_4kg_lift_up}, respectively. Payloads heavier than this were avoided due to the towing mechanism for tendon drive and safety concerns. The angle of the joint is such that the amplitude is $45^{\circ}$ under load, we manually sent the command values from the game controller.

\subsection{Simple Manipulation}
\begin{figure}
    \centering
    \includegraphics[width=0.48\textwidth]{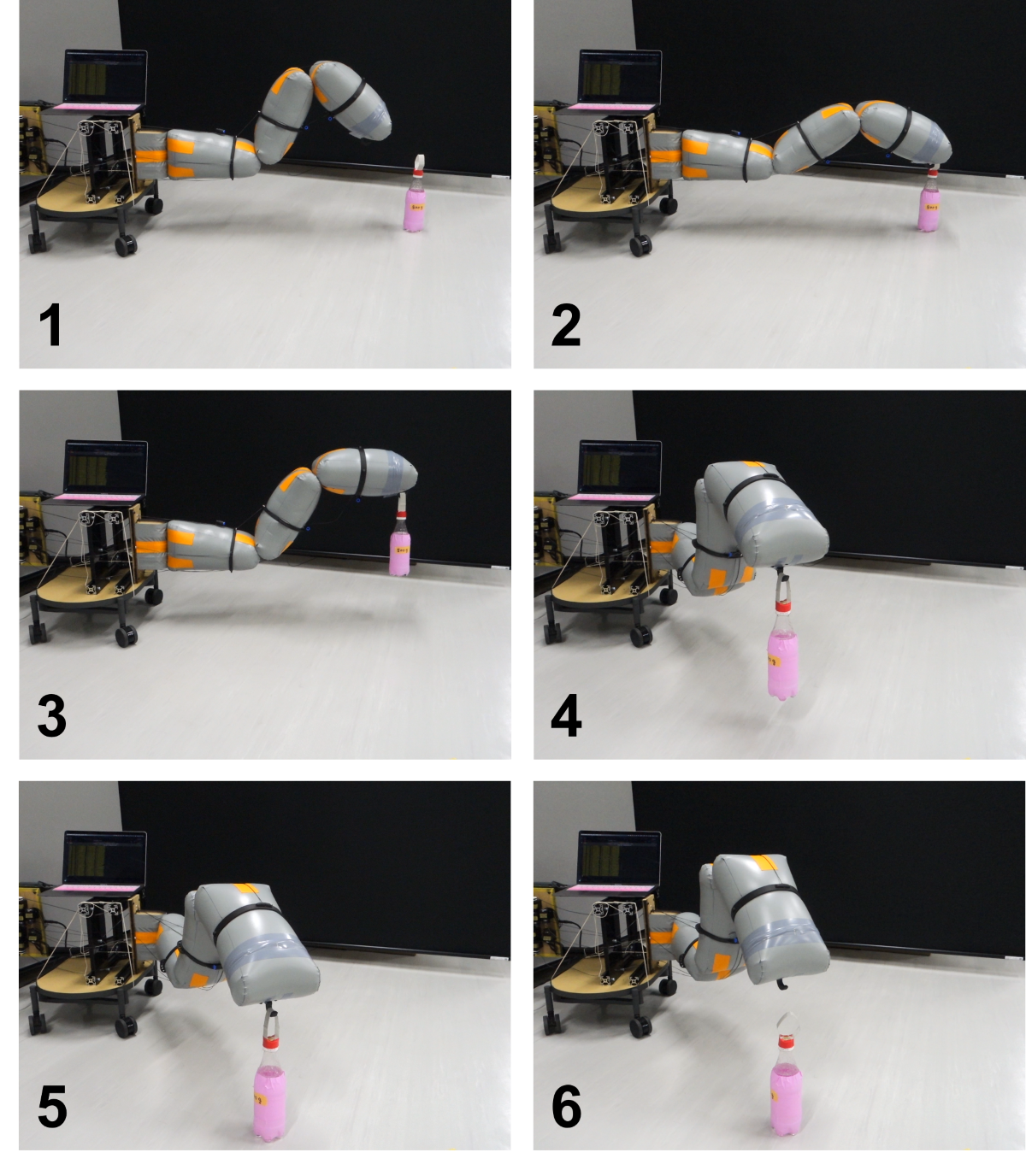}
    \caption{Manipulation experiment with 500 g payload. About a minute from start to finish.}
    \label{fig:demo_3dof_pick_up}
\end{figure}

We attached a hook to the end of a 3-DoF robotic arm to pick and place the object's movement (Fig.\ref{fig:demo_3dof_pick_up}). A 500~g payload was hooked and released at a distance. The release operation consisted of placing the hooked payload on the floor and unhooking it when it floated off the hook. This manipulation was performed by sending commands for end effector position by a person using a game controller and solving inverse kinematics.

After the load of the payload was applied to the arm during the lifting operation, it did not lift for a while even though the servo motor for tendon driven was rotating. This means that the difference from the target angle would be noticeable depending on the presence or absence of the payload. The flexibility of the pneumatic bladder link as well as the flexibility of the tendon drive amplified the difference from the target.

\subsection{Locomotion}
\begin{figure}
    \centering
    \includegraphics[width=0.48\textwidth]{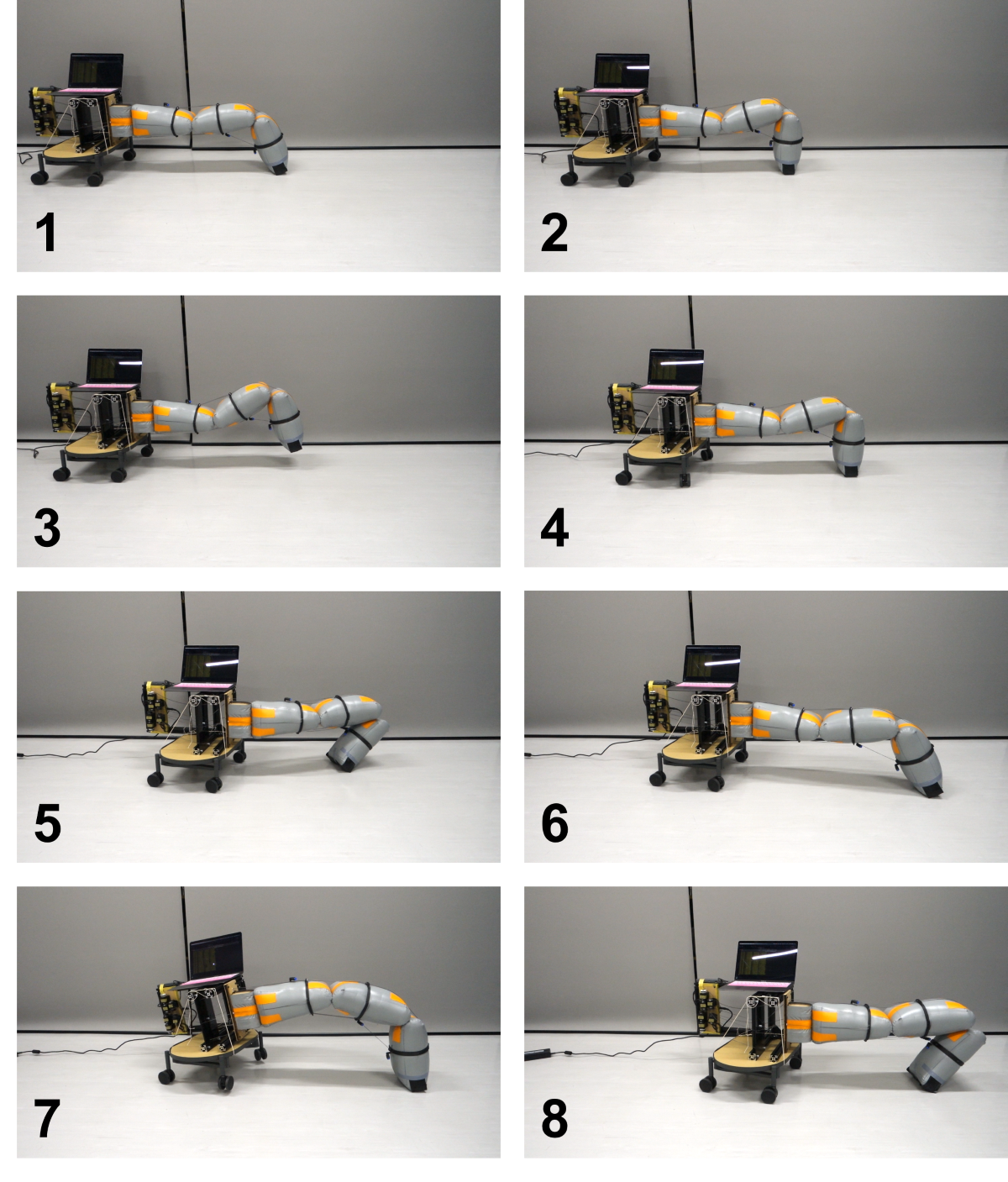}
    \caption{Single leg locomotion experiment.  About 1.5 minutes from start to finish. The distance traveled is just over 1 meter.}
    \label{fig:demo_3dof_locomotion}
\end{figure}

We mounted a 3-DoF robotic arm on a dolly and performed locomotion as a 3-DoF single leg (Fig.\ref{fig:demo_3dof_locomotion}). The trajectory of the leg tips was manipulated by a human using a game controller. In other words, the pilot used his senses to control the leg tips to form a gait using his vision. Even with a somewhat crude trajectory, the load and errors were absorbed from the compliance inherent in inflatable robots. In this experiment, a rubber sheet was attached to the tip of the robot arm because the frictional force between the tip of the arm and the floor surface was not sufficient.

\section{Discussion}
The wide range of motion of the proposed inflatable robot provided sufficient working space. The robots lifted a payload up to 5 kg, performed simple manipulation and single-leg locomotion, although they were operated by a human operator. The combination of the pneumatic bladder link and the Hilberry joints was fabricated and its effectiveness was shown. By making a combination of these parts according to purpose, an inexpensive and lightweight robot can be fabricated.

The pull length of tendon-driven was referenced by applying tension in the initial posture (in this case, all joint angles were $0^{\circ}$ at initial), and the difference from the expected string length at the estimated angle was used as the target value. However, due to the softness of the pneumatic bladder link, the distance between anchor points changed depending on the condition. Therefore, a large discrepancy occurred between the assumed angle and the actual angle. This indicates that it is very difficult to position an object accurately when it is grasped in manipulation. In this experiment, the experimenter visually provided feedback in all experiments. Therefore, this characteristic did not have a significant impact. However, this will be an issue when the system is operated as a stand-alone robot system. It will be necessary to implement sensing or estimation of deflection according to the weight of the payload based on the internal pressure of the link and the arm's posture. In walking locomotion, the load from the ground in the stance phase and the no-load state in the swing phase keep alternating in a single task of locomotion. In any case, more detailed state recognition is necessary to enhance the application.

\section{Conclusion}
In this work, we proposed an inflatable robot consisting of a pneumatic bladder links and Hilberry joints. The characteristics of the adopted joints enable a wide range of motion of $\pm150^{\circ}$, which is the largest for an inflatable robot. In addition, the design of one pneumatic bladder per link for future modularization is expected to realize a robot arm with more degrees of freedom. The actual manipulation and locomotion tasks performed on the 3-DoF robot arm demonstrated the potential for real-world applications of inflatable robots. As more degrees of freedom and applications are added, the future focus will be on sensing and other methods to provide feedback on the state of the robot.

\section*{Acknowledgment}
This work was supported by JST Moonshot R\&D Grant Number JPMJMS2013.
\bibliography{inflatable}

@patent{Hilberry1973:Rolling_Contact_Joint,
author       = {Benny M. Hillberry and Allen S. Hall, Jr.},
  title        = {Rolling contact joint},
  nationality = "United States",
  number = "3932045",
  day = "5",
  month = Mar,
  year = "1973",
  url = "https://google.com/patents/US3932045A"
}

@inproceedings{Sanan2011:Inflatable_Robot_Arm,
  author    = {Sanan, Siddharth and Ornstein, Michael H. and Atkeson, Christopher G.},
  booktitle = {2011 Annual International Conference of the IEEE Engineering in Medicine and Biology Society},
  title     = {Physical human interaction for an inflatable manipulator},
  year      = {2011},
  volume    = {},
  number    = {},
  pages     = {7401-7404},
  doi       = {10.1109/IEMBS.2011.6091723}
}

@article{Takeichi2017:Giacometti_Arm,
  author  = {Takeichi, M. and Suzumori, K. and Endo, G. and Nabae, H.},
  journal = {IEEE Robotics and Automation Letters},
  title   = {Development of Giacometti Arm With Balloon Body},
  year    = {2017},
  volume  = {2},
  number  = {2},
  pages   = {951-957},
  doi     = {10.1109/LRA.2017.2655111}
}

@inproceedings{Takeichi2017:20m_Giacometti_Arm,
  author    = {Takeichi, Masashi and Suzumori, Koichi and Endo, Gen and Nabae, Hiroyuki},
  booktitle = {2017 IEEE/RSJ International Conference on Intelligent Robots and Systems (IROS)},
  title     = {Development of a 20-m-long Giacometti arm with balloon body based on kinematic model with air resistance},
  year      = {2017},
  volume    = {},
  number    = {},
  pages     = {2710-2716},
  doi       = {10.1109/IROS.2017.8206097}
}

@inproceedings{Bodily2017:Multi_objective_Design_Optimization,
  author    = {Bodily, Daniel M. and Allen, Thomas F. and Killpack, Marc D.},
  booktitle = {2017 IEEE International Conference on Robotics and Automation (ICRA)},
  title     = {Multi-objective design optimization of a soft, pneumatic robot},
  year      = {2017},
  volume    = {},
  number    = {},
  pages     = {1864-1871},
  doi       = {10.1109/ICRA.2017.7989218}
}

@article{Qi2017:Safe_Inflatable_Robot_for_HRI,
  author={Qi, Ronghuai and Khajepour, Amir and Melek, William W. and Lam, Tin Lun and Xu, Yangsheng},
  journal={IEEE Transactions on Robotics}, 
  title={Design, Kinematics, and Control of a Multijoint Soft Inflatable Arm for Human-Safe Interaction}, 
  year={2017},
  volume={33},
  number={3},
  pages={594-609},
  doi={10.1109/TRO.2016.2647231}
}

@article{Liang2018:Soft_Robotic_Arm,
  author  = {Liang, Xinquan and Cheong, Haris and Sun, Yi and Guo, Jin and Chui, Chee Kong and Yeow, Chen-Hua},
  journal = {IEEE Robotics and Automation Letters},
  title   = {Design, Characterization, and Implementation of a Two-DOF Fabric-Based Soft Robotic Arm},
  year    = {2018},
  volume  = {3},
  number  = {3},
  pages   = {2702-2709},
  doi     = {10.1109/LRA.2018.2831723}
}

@article{Kim2018:High_Accuracy_Inflatable_Robot,
  author = {Hye-Jong Kim, Akihiro Kawamura, Yasutaka Nishioka and Sadao Kawamura},
  title = {Mechanical design and control of inflatable robotic arms for high positioning accuracy},
  journal = {Advanced Robotics},
  volume = {32},
  pages = {89-104},
  year = {2018},
  publisher = {Taylor & Francis},
  doi = {10.1080/01691864.2017.1405845},
  URL = {https://doi.org/10.1080/01691864.2017.1405845},
  eprint = {https://doi.org/10.1080/01691864.2017.1405845}
}

@inproceedings{Seong2019:Low_pressure_Soft_Inflatable_Joint,
  author    = {Seong, Young Ah and Niiyama, Ryuma and Kawahara, Yoshihiro and Kuniyoshi, Yasuo},
  booktitle = {2019 2nd IEEE International Conference on Soft Robotics (RoboSoft)},
  title     = {Low-pressure Soft Inflatable Joint Driven by Inner Tendon},
  year      = {2019},
  volume    = {},
  number    = {},
  pages     = {37-42},
  doi       = {10.1109/ROBOSOFT.2019.8722764}
}

@article{Hyatt2019:Configuration_Estimation_IRA,
  author  = {Hyatt, Phillip and Kraus, Dustan and Sherrod, Vallan and Rupert, Levi and Day, Nathan and Killpack, Marc D.},
  journal = {IEEE/ASME Transactions on Mechatronics},
  title   = {Configuration Estimation for Accurate Position Control of Large-Scale Soft Robots},
  year    = {2019},
  volume  = {24},
  number  = {1},
  pages   = {88-99},
  doi     = {10.1109/TMECH.2018.2878228}
}

@article{Li2020:Design_Modeling_Characterization_IRA,
  title    = {Design, modeling and characterization of a joint for inflatable robotic arms},
  journal  = {Mechatronics},
  volume   = {65},
  pages    = {102311},
  year     = {2020},
  issn     = {0957-4158},
  doi      = {https://doi.org/10.1016/j.mechatronics.2019.102311},
  url      = {https://www.sciencedirect.com/science/article/pii/S0957415819301436},
  author   = {XueAi Li and Kui Sun and Chuangqiang Guo and Teng Liu and Hong Liu}
}

@article{Li2023:Large_Scale_IRA,
  author  = {Li, XueAi and Yue, Honghao and Yang, Dapeng and Sun, Kui and Liu, Hong},
  journal = {IEEE Transactions on Industrial Electronics},
  title   = {A Large-Scale Inflatable Robotic Arm Toward Inspecting Sensitive Environments: Design and Performance Evaluation},
  year    = {2023},
  volume  = {70},
  number  = {12},
  pages   = {12486-12499},
  doi     = {10.1109/TIE.2022.3232643}
}

@article{Oh2023:Large_Scale_Inflatable_Robotic_Arms,
  title   = {Toward the Development of {Large-Scale} Inflatable Robotic Arms Using Hot Air Welding},
  author  = {Oh, Namsoo and Rodrigue, Hugo},
  journal = {Soft Robotics},
  volume  = 10,
  number  = 1,
  pages   = {88-96},
  month   = feb,
  year    = 2023,
  doi     = {10.1089/soro.2021.0134}
}

\end{document}